\documentclass[preprint,
superscriptaddress,
amsmath,amssymb,aps,showkeys,showpacs,
twoside,final,secnumarabic,%
nofootinbib]{revtex4-2}

\usepackage[a4paper,margin=20mm]{geometry}

\usepackage{cmap}
\usepackage[T1,T2A]{fontenc}
\usepackage[utf8]{inputenc}
\usepackage[russian,english]{babel}
\usepackage{color}
\usepackage{graphicx}%
\usepackage{dcolumn}%
\usepackage{bm}
\usepackage[unicode=true,colorlinks=true,linkcolor=magenta, urlcolor=blue, citecolor = blue,breaklinks]{hyperref}
\usepackage{multirow}
\usepackage{float}
\usepackage{longtable}
\usepackage{url}
\usepackage{breakurl}
\DeclareGraphicsExtensions{.eps,.png,.pdf}

\begin{document}

\title{Feature-Based and Deep Learning Methods for Astronomical Light Curve Classification}

\def\addressa{HSE University}
\def\addressb{Southern Federal University}
\def\addressc{Lomonosov Moscow State University}
\def\addresse{Max-Planck-Institut f\"ur Astronomie}
\def\addressf{Heidelberg University}
\def\addressg{Carnegie Mellon University}

\author{\firstname{V.}~\surname{Baymakanov}}
\email[E-mail: ]{vkbaymakanov@edu.hse.ru}
\affiliation{\addressa}
\author{\firstname{Kh.}~\surname{Khamitov}}
\affiliation{\addressa}
\author{\firstname{A.}~\surname{Raimova}}
\affiliation{\addressa}
\author{\firstname{A.}~\surname{Isaeva}}
\affiliation{\addressb}
\author{\firstname{R.}~\surname{Taran}}
\affiliation{\addressb}
\author{\firstname{M.}~\surname{Zedan}}
\affiliation{\addressa}
\author{\firstname{M.}~\surname{Kornilov}}
\affiliation{\addressa}
\affiliation{\addressc}
\author{\firstname{M.}~\surname{Demianenko}}
\affiliation{\addresse}
\affiliation{\addressf}
\author{\firstname{K.}~\surname{Malanchev}}
\affiliation{\addressg}
\author{\firstname{M.}~\surname{Hushchyn}}
\affiliation{\addressa}

\begin{abstract}

We compare feature-based machine learning and end-to-end deep learning for variable star classification using the Zwicky Transient Facility (ZTF) StarEmbed dataset, which contains about 40,000 objects across seven imbalanced classes. Classical models trained on physically motivated features consistently outperform deep networks trained directly on raw light curves. We find that limited labeled data contributes to the performance gap, while the multivariate, irregularly sampled, and morphologically diverse nature of light curves poses additional challenges for deep learning models. While deep learning holds promise for learning representations directly from observations, our results indicate that current models fail to generalize with the available training set size. These findings motivate further investigation of the benefits of scaling labeled datasets and self-supervised pretraining on large unlabeled surveys. This work establishes a quantitative baseline for future research and highlights the need for further innovations to unlock the potential of deep learning in astronomical time series analysis.

\end{abstract}

\par
\keywords{machine learning, deep learning, light curves, variable stars, ZTF \\[5pt]}

\maketitle

\noindent\textit{Submitted to Moscow University Physics Bulletin (special issue ``Deep Learning in Computational Physics'', DLCP-2026).}

\section{Introduction}\label{intro}

Modern astronomical sky surveys, such as the Zwicky Transient Facility (ZTF)~\cite{1} and the Legacy Survey of Space and Time (LSST)~\cite{2}, collect enormous amounts of data. For example, the Vera C.~Rubin Observatory~\cite{2}, which recently began its operation, generates tens of terabytes of data per day, and this volume will only increase with the commissioning of new-generation telescopes. Such data cannot be processed manually within a reasonable time frame, yet their processing is essential. Astronomers rely on it for a range of tasks, from classifying astronomical objects and detecting transient events to determining their astrophysical properties. Machine learning and deep learning methods enable the development of models that can efficiently handle these continuously growing data volumes.

The objects studied are usually divided into three broad categories: transients, such as supernova explosions~\cite{3,4}; periodic objects, whose brightness variations repeat with a certain period~\cite{5}; and aperiodic or stochastic objects, whose brightness variations are noise-like~\cite{6,7}.

In this work, we address the multiclass classification of astronomical objects by their light curves, which is one of the most common and important tasks in modern astronomy. A light curve describes the variation of an object's flux over time in one or more photometric filters, often referred to as bands. It can be viewed as a multivariate time series that typically exhibits an irregular observation cadence and missing data points due to observational constraints~\cite{8}. Moreover, each measurement comes with an associated uncertainty. These factors collectively make light-curve analysis challenging and should be taken into account for training machine learning models.

The classical approaches, developed over the past decades, build a separate specialized model for each task and dataset~\cite{9,10}. For each light curve, several physically motivated features are extracted, and a machine-learning classifier is then fitted to these features. This approach is straightforward, transparent, and can yield good results, but models built this way transfer poorly to new surveys, with performance often degrading across datasets.

A more recent, promising alternative is based on foundation models~\cite{11,12}, trained in a self-supervised regime to learn meaningful data representations, which are then fed into task-specific models. These models are universal and can be applied to any dataset and downstream task. However, their classification quality still underperforms compared to classical approaches. The quality gap between these two paradigms has not been fully investigated in previous works.

In this work, we study multiclass classification of astronomical objects by their light curves and compare two approaches: training classical machine learning models, such as Logistic Regression, CatBoost, and Multilayer Perceptron (MLP), on pre-extracted physically motivated features, and training deep neural networks with Transformer and Long Short-Term Memory (LSTM) architectures directly on raw observational data.

We test the hypothesis that deep learning networks can learn meaningful representations directly from raw light curves and are capable of solving the classification task. We also aim to quantify the performance gap between these two paradigms, identify the factors underlying their differences, and evaluate whether the observed discrepancy stems from architectural limitations or from the intrinsic complexity of the data, including its irregular sampling, multivariate nature, and morphological diversity. In particular, we investigate the extent to which physically motivated augmentations, operating exclusively on real observations, can mitigate the data limitation of deep architectures and narrow the performance gap.

\section{Dataset}

We use the ZTF StarEmbed dataset as the benchmark throughout our study. This dataset is a subset of the ZTF astronomical survey~\cite{1}, conducted at the Palomar Observatory, and contains light curves of stars belonging to seven distinct classes: EW and EA, which are eclipsing binaries; RRab, RRc, and RRd, which are subtypes of RR~Lyrae pulsating stars; RS~CVn, which are binary star systems; and LPV, which are long-period variable stars~\cite{13}. Originally introduced in \cite{14}, the dataset was designed as a benchmark for comparing different classifiers. It comprises roughly 40,000 objects and about 35 million individual observations, with a median light-curve length of approximately $770$ data points.

The light curves are recorded in three ZTF photometric bands, namely $g$, $r$, and $i$. Each observation is represented as a tuple $(t, \mathrm{mag}, \mathrm{mag\_err}, \mathrm{passband})$, where $t$ denotes the observation time in Modified Julian Date, $\mathrm{mag}$ is the stellar magnitude, $\mathrm{mag\_err}$ is the associated measurement uncertainty, and $\mathrm{passband}$ specifies the filter used. The dataset is stored in the nested-pandas format, in which each light curve is organized as a nested table within a row of the main DataFrame. This structure enables efficient vectorized operations across the entire dataset. We split the data into training, validation, and test sets in a 7:1:2 ratio, while preserving the class proportions in each subset.

\begin{figure}[t]
\centering
\includegraphics[width=1.\linewidth]{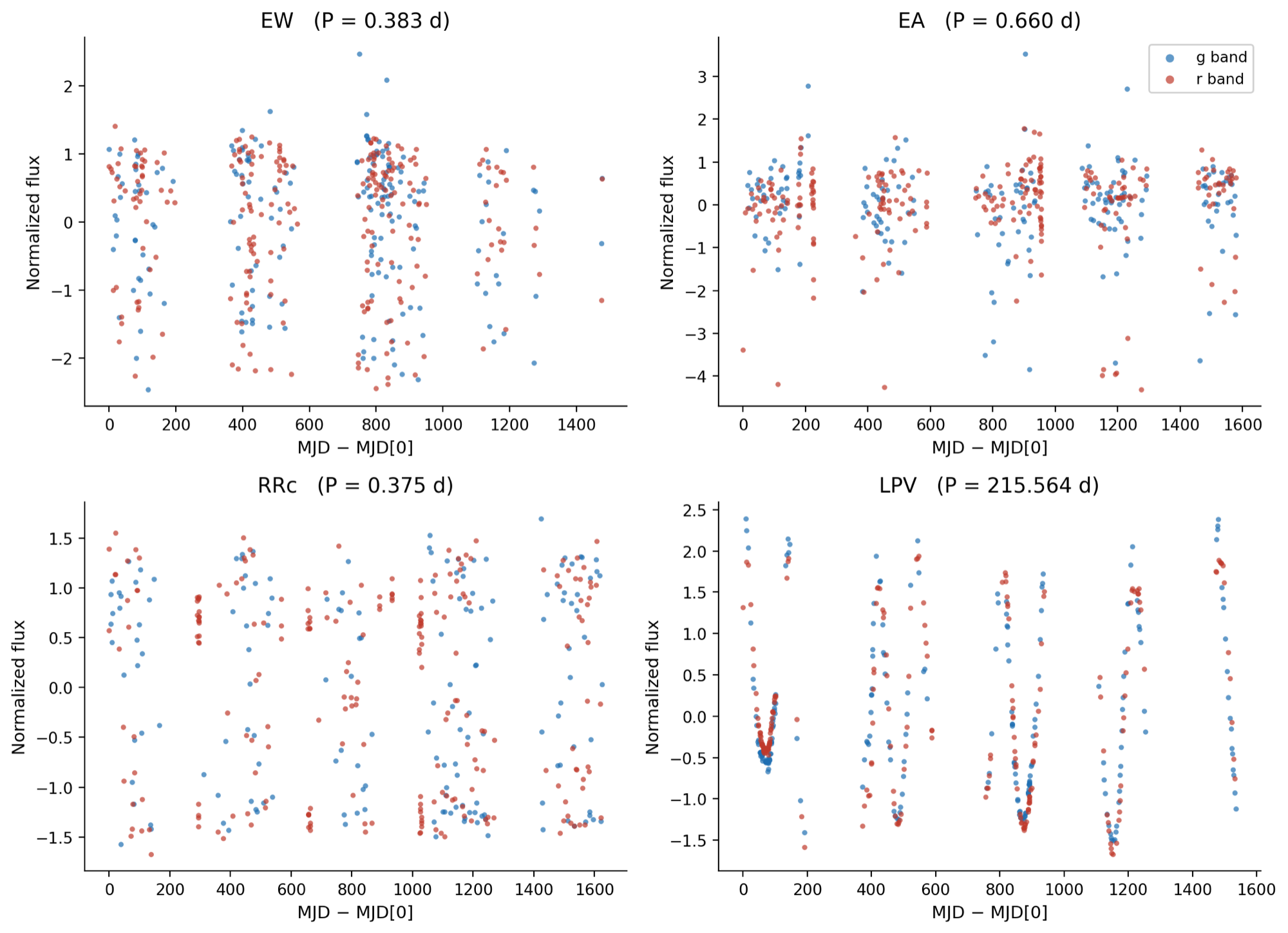}
\caption{\label{fig:lightcurves} Examples of light curves from the ZTF StarEmbed dataset for the EW, EA, RRc, and LPV classes in the $g$ and $r$ passbands.}
\end{figure}

The dataset suffers from a strong class imbalance, with the number of objects per class detailed in Table~\ref{tab:classes}. This imbalance poses additional challenges for model training and necessitates a careful selection of evaluation metrics. Representative light curves from each class are shown in Fig.~\ref{fig:lightcurves}.

\begin{table}[htb]
\caption{\label{tab:classes} Number of objects per class in the ZTF StarEmbed dataset.}
\renewcommand{\arraystretch}{0.62}
\begin{ruledtabular}
\begin{tabular}{lrrrrrrrr}
Class & EW & RRc & EA & RRab & RS~CVn & LPV & RRd & Total \\
\hline
Number of objects & 27{,}949 & 4{,}792 & 4{,}251 & 2{,}068 & 1{,}401 & 457 & 439 & 41{,}357 \\
\end{tabular}
\end{ruledtabular}
\end{table}

\section{Evaluation Metrics}

To evaluate model performance and enable fair comparisons, we employ four metrics: accuracy, balanced accuracy, macro-averaged F1, and the macro-averaged area under the receiver operating characteristic curve (ROC-AUC).

The choice of these metrics is primarily driven by the class imbalance present in the dataset. The largest class, EW, contains roughly 64 times more objects than the smallest class, RRd. Under such conditions, plain accuracy becomes biased toward the majority classes, as a classifier that disregards rare classes can still achieve a high accuracy score. For this reason, we report accuracy only as a baseline reference. Our primary evaluation relies on macro-averaged metrics, which treat all classes equally regardless of their size. This ensures that performance on rare classes such as LPV and RRd contributes to the final score on the same footing as performance on the dominant EW class.

\section{Feature-Based Approach}

We first explored an approach based on physically informative features extracted from light curves. For each photometric passband, we computed 25 features using the \texttt{light-curve-python} library~\cite{15}. These features were then concatenated and used to train three classifiers: Logistic Regression in the one-vs-rest regime, CatBoost~\cite{16}, and a Multilayer Perceptron.

\begin{figure}[t]
\centering
\includegraphics[width=0.5\linewidth]{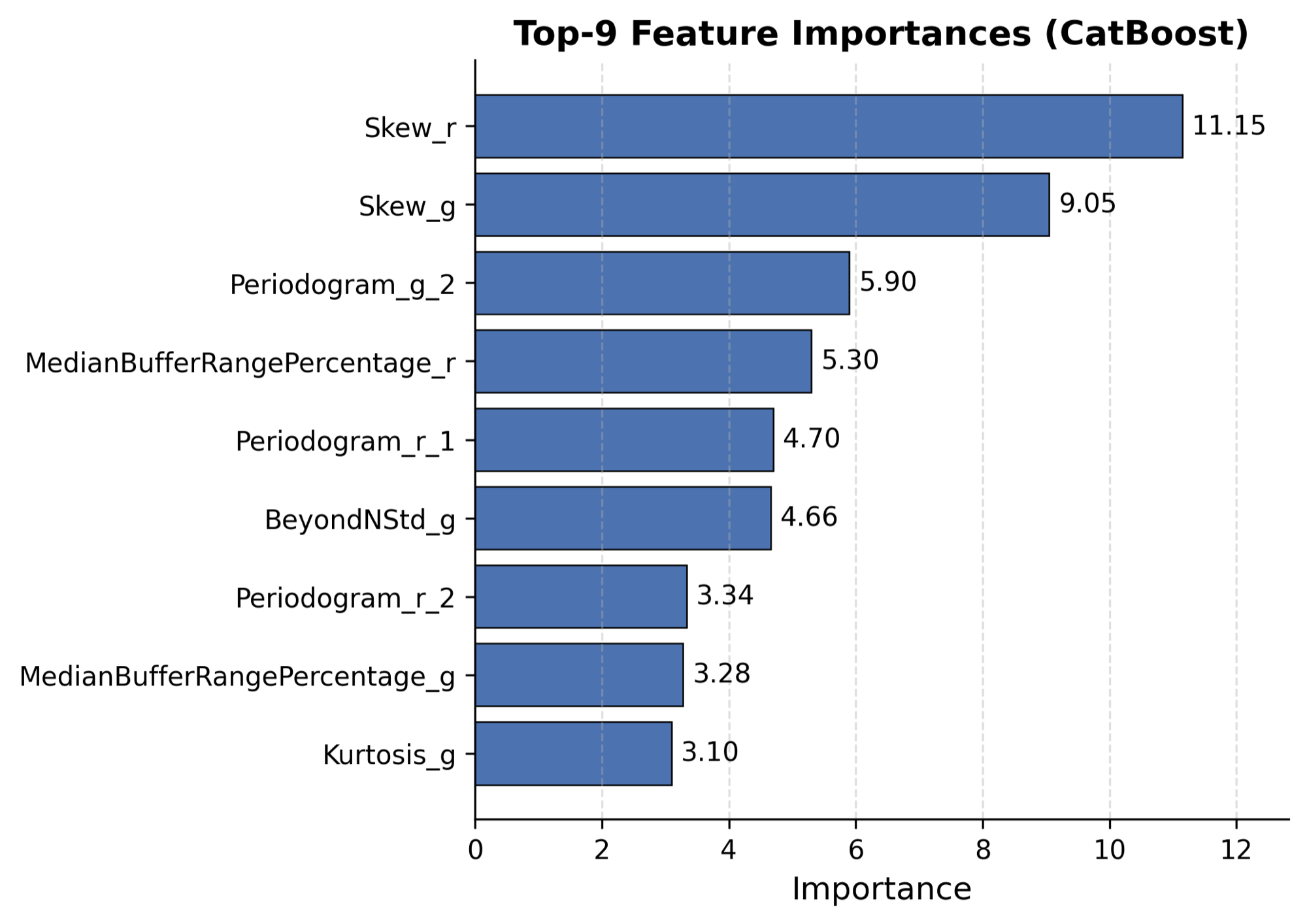}
\caption{\label{fig:feature_importance} Top-9 features with the highest importance scores, as determined by the CatBoost classifier trained on light-curve features.}
\end{figure}

For each light curve, we independently computed a set of 25 statistical and periodogram-based features for each photometric band. The full list of features includes Amplitude, AndersonDarlingNormal, BeyondNStd, Cusum, Eta, EtaE, ExcessVariance, InterPercentileRange, Kurtosis, MaximumSlope, Mean, MeanVariance, Median, MedianAbsoluteDeviation, MedianBufferRangePercentage, PercentAmplitude, PercentDifferenceMagnitudePercentile, ReducedChi2, Roms, Skew, StandardDeviation, StetsonK, WeightedMean, and Periodogram, which yields two outputs. Detailed descriptions of these features are available on the \texttt{light-curve-python} library website\footnote{\url{https://light-curve.snad.space/latest/features/}}. Figure~\ref{fig:feature_importance} presents the nine features identified as the most important for classification by the CatBoost model.

\section{Deep Learning Approach}

The second approach relies on training deep neural networks directly on raw light curves, omitting a separate feature extraction stage. Both architectures considered in this work share a common input representation. Each light curve is converted into a sequence of up to 512 observations, with each observation described by four features: the time in days since the first observation, the logarithm of the passband wavelength, the normalized flux, and the normalized flux uncertainty, the latter obtained by dividing the measurement uncertainty by the standard deviation of the flux. Light curves shorter than 512 observations are zero-padded, and a corresponding padding mask is passed to the network to exclude padded positions from computations.

Both networks are trained using the AdamW optimizer with a class-weighted cross-entropy loss, which helps mitigate the severe class imbalance in the dataset. Each of the configurations was additionally trained with each of the three light-curve augmentation strategies, described in detail in Section~\ref{sec:augmentations}.

\subsection{LSTM}

For the recurrent approach, we explored both unidirectional and bidirectional networks with varying depths and classification heads. Bidirectional processing is appropriate in this setting because classification considers the entire light curve, so information from future time steps does not violate the problem formulation. The simplest configuration we tested consists of a single unidirectional LSTM~\cite{17} layer with a hidden size of 128, whose final hidden state is passed to a linear classifier. In deeper configurations, we aggregate the hidden states produced by the network using multiple pooling strategies: the mean, maximum, and standard deviation over all time steps, together with the final hidden state vectors. These aggregated representations are concatenated and fed into the classification head. Several head variants were considered, including a head with Gaussian Error Linear Unit (GELU) activation, a wide head, and a combination of both.

\begin{figure}[t]
\centering
\includegraphics[width=0.5\linewidth]{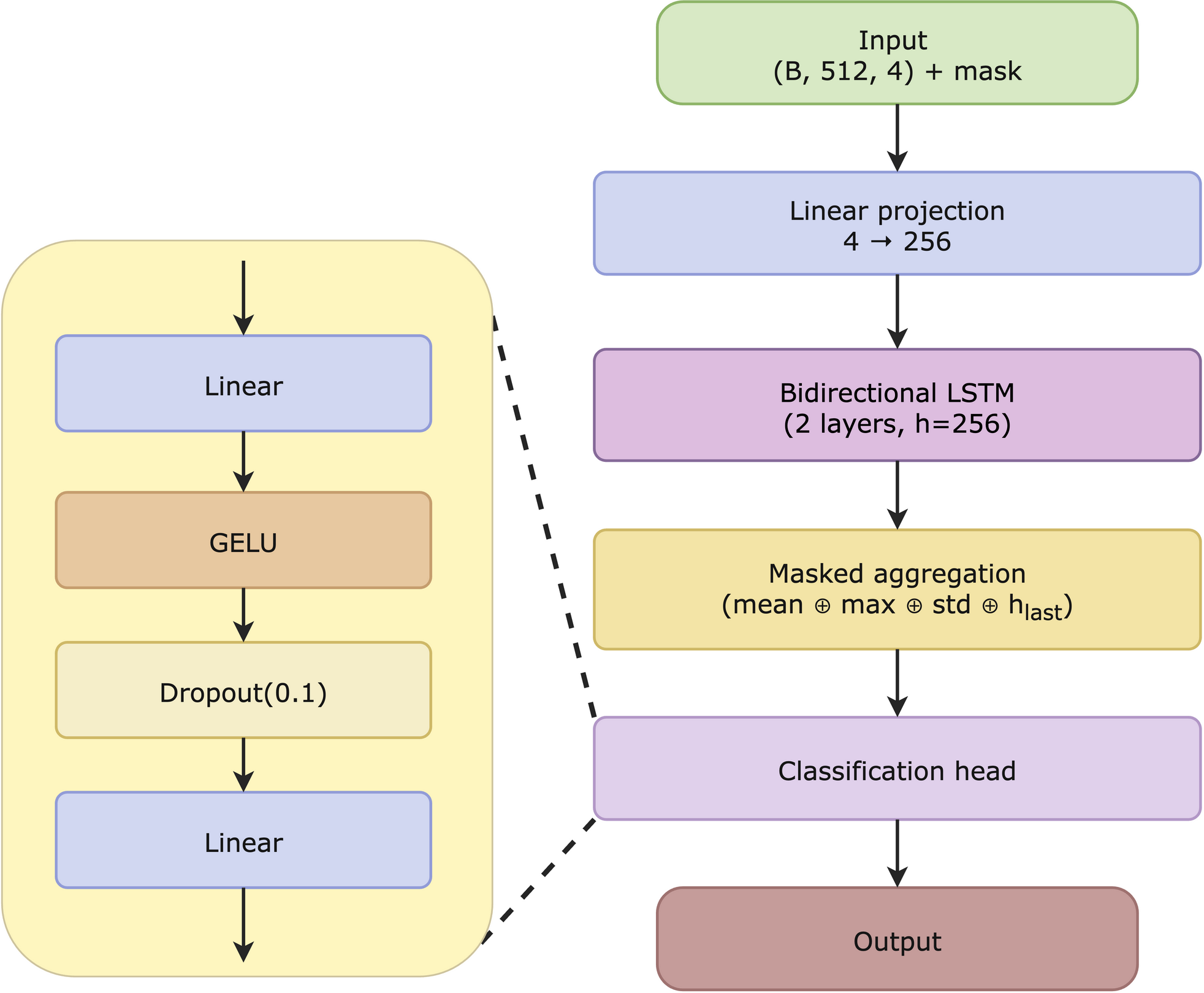}
\caption{\label{fig:lstm} The best-performing recurrent architecture: input projection, two bidirectional LSTM layers with hidden size 256, masked aggregation, and a classification head.}
\end{figure}

The best-performing architecture is illustrated in Figure~\ref{fig:lstm}. It comprises two bidirectional LSTM (BiLSTM) layers with a hidden size of 256, followed by masked aggregation combining the mean, maximum, standard deviation, and final state, and a classification head. The model contains approximately 3.7 million parameters. Training was regularized with early stopping on the validation set.

\subsection{Transformer}

\begin{figure}[t]
\centering
\includegraphics[width=0.5\linewidth]{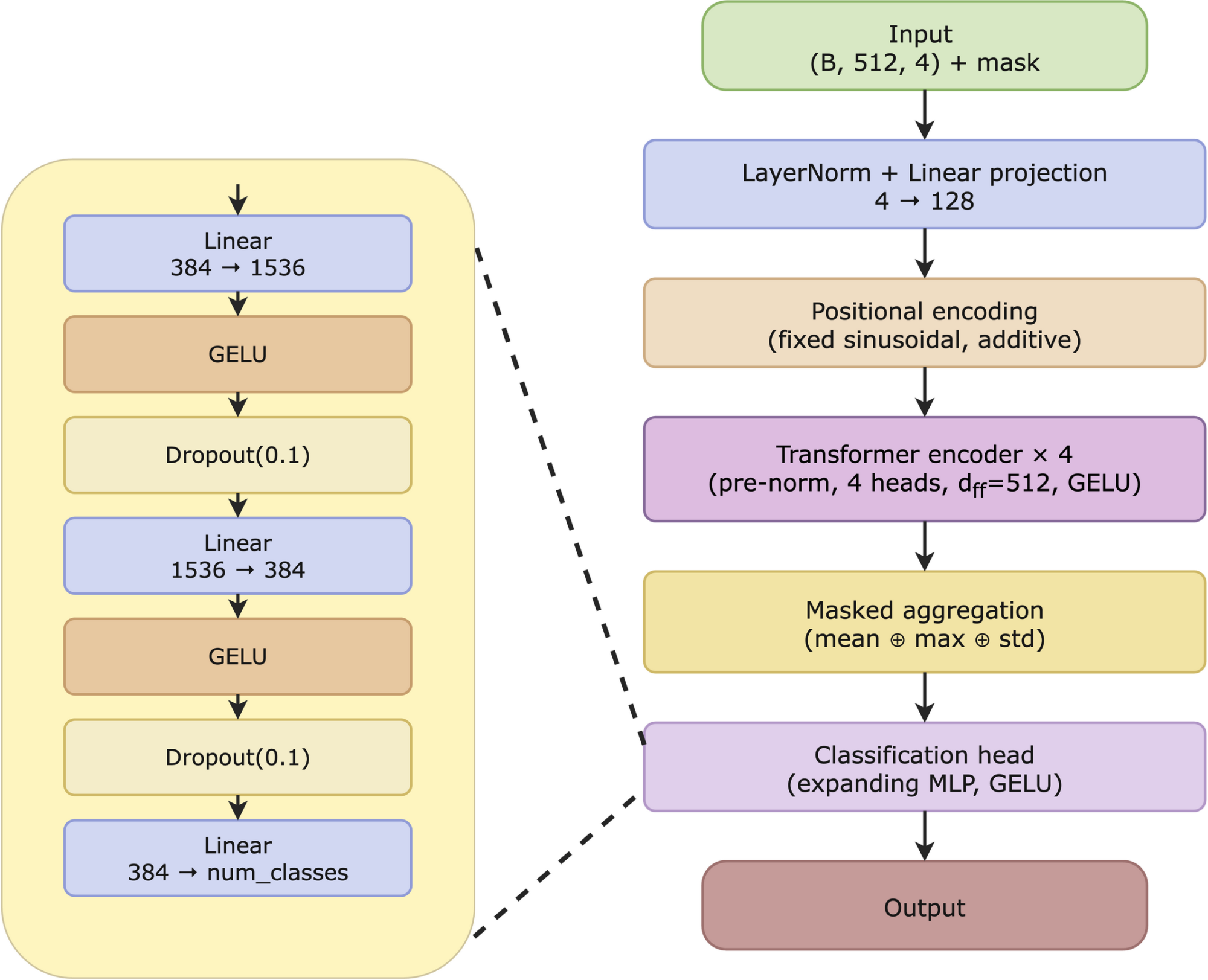}
\caption{\label{fig:transformer} The deep Transformer architecture: input projection, sinusoidal positional encoding, four pre-norm encoder layers, masked aggregation, and an MLP classification head.}
\end{figure}

We also examined a Transformer~\cite{18} encoder applied to the same input sequences. The encoder is configured with a model dimension of 128, four attention heads, a feed-forward dimension of 512, a pre-normalization layer arrangement, and fixed sinusoidal positional encodings added to the input projection. The simplest configuration consists of two encoder layers with mean pooling and a linear classification head, containing only about 0.27 million parameters. The deeper configuration, shown in Figure~\ref{fig:transformer}, employs four encoder layers and aggregates the encoder outputs via masked mean, maximum, and standard deviation pooling, followed by an expanding MLP head with GELU activation. During architecture selection, we compared narrow and wide head variants with both Rectified Linear Unit (ReLU) and GELU activations.

The Transformer was trained with a learning-rate warmup followed by a cosine decay schedule.
\vspace{\baselineskip}

\subsection{Light-Curve Augmentations}\label{sec:augmentations}

\begin{figure}[t]
\centering
\includegraphics[width=\linewidth]{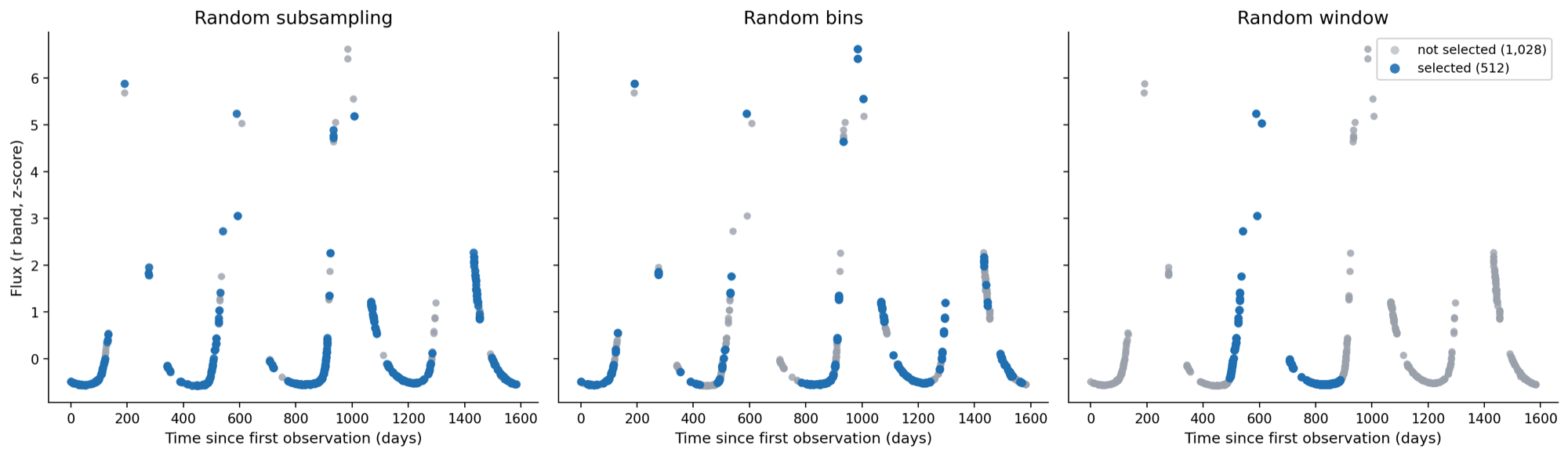}
\caption{\label{fig:augmentations} The three augmentations applied to the same light curve of 2,982 observations; the 512 selected observations are shown in blue, the rest in gray.}
\end{figure}

Since the input length is limited to 512 observations while the median light curve in the dataset contains about 770, most light curves cannot be fed to the networks in full, and some subset of observations must be selected in any case. This constraint makes augmentation a natural choice in our setting. Instead of fixing a single deterministic truncation, we randomly draw the subset at each training epoch. No artificial data points are generated. Every augmented light curve consists solely of real observations of the same object and preserves the irregular sampling inherent to the survey. The augmentations therefore do not distort the physical content of the data while substantially increasing its effective diversity. Three augmentation strategies were considered and are illustrated in Figure~\ref{fig:augmentations}.

\textit{Random subsampling.} 512 observations are drawn uniformly at random from the entire light curve and kept in chronological order. This preserves the full time span of the curve at the cost of a sparser cadence.

\textit{Random bins.} 24-hour bins are selected at random, and all observations within each chosen bin are retained until 512 points are collected; any excess points from the final bin are discarded. This strategy preserves the dense intra-night structure of the observations.

\textit{Random window.} 512 consecutive observations are kept starting from a random position, which emulates a shorter observing campaign.

\section{Model Comparison}

Table~\ref{tab:results} presents a comprehensive comparison of all evaluated models on the held-out test set, encompassing both feature-based classical machine learning approaches and deep neural networks trained directly on raw light curve observations. The results reveal a clear and consistent superiority of the feature-based pipeline across virtually all evaluation metrics, underscoring the continued value of domain-specific feature engineering in astronomical classification tasks.

Among the feature-based models, CatBoost achieves the highest accuracy of 0.907 and, together with the MLP, the top macro-averaged ROC-AUC of 0.981. The MLP performs comparably with 0.903 accuracy, indicating that the extracted features form a rich and nearly linearly separable representation. Interestingly, logistic regression attains the highest balanced accuracy of 0.801, outperforming both CatBoost and the MLP. We attribute this to its linear nature, which reduces overfitting to majority classes and better preserves performance across the imbalanced class distribution.

Deep learning models trained from scratch on raw light curves exhibit substantially inferior performance relative to feature-based approaches. The best unaugmented configuration, the LSTM, achieves only 0.609 accuracy, 0.508 macro-averaged F1, and 0.914 macro-averaged ROC-AUC. This substantial performance gap underscores the inherent difficulty of learning robust representations directly from irregularly sampled, multivariate time series, corroborating our hypothesis that deep architectures are data-limited in this regime.

However, physically motivated augmentations considerably reduce this performance deficit. For the LSTM-like models, random windows and random daily bins emerge as the most effective augmentation strategies across the different configurations. In particular, the best-performing configuration raises accuracy from 0.609 to 0.763, macro-averaged F1 from 0.508 to 0.625, and macro-averaged ROC-AUC from 0.914 to 0.948.

For the Transformer-like models, random daily bins gives the best results in most configurations. Overall, the best-performing configuration improves accuracy from 0.588 to 0.692, macro-averaged F1 from 0.488 to 0.556, and macro-averaged ROC-AUC from 0.894 to 0.923. 

Both random bins and random window improve balanced accuracy in nearly all configurations, whereas random subsampling is less stable and degrades performance in several cases. Which of the two strategies performs best varies from configuration to configuration without a consistent pattern, and without further investigation these differences cannot be separated from run-to-run variability.

\begin{table}[tbh]
\renewcommand{\arraystretch}{0.62}
\caption{\label{tab:results} Test-set results of all experiments (standard error as subscript). Best values in bold.}
\begin{ruledtabular}
\begin{tabular}{lcccc}
Architecture & Acc. & Bal.\ Acc. & Macro F1 & ROC-AUC \\
\hline
Extracted Features (Logistic Regression) & $0.787_{\scriptstyle .005}$ & $\textbf{0.801}_{\scriptstyle .011}$ & $0.661_{\scriptstyle .008}$ & $0.955_{\scriptstyle .005}$ \\
Extracted Features (CatBoost) & $\textbf{0.907}_{\scriptstyle .004}$ & $0.742_{\scriptstyle .009}$ & $0.771_{\scriptstyle .009}$ & $\textbf{0.981}_{\scriptstyle .003}$ \\
Extracted Features (MLP) & $0.903_{\scriptstyle .003}$ & $0.762_{\scriptstyle .010}$ & $\textbf{0.779}_{\scriptstyle .009}$ & $\textbf{0.981}_{\scriptstyle .004}$ \\
\hline
LSTM (Linear) & $0.609_{\scriptstyle .005}$ & $0.653_{\scriptstyle .011}$ & $0.508_{\scriptstyle .008}$ & $0.914_{\scriptstyle .003}$ \\
LSTM (Linear), Rnd.\ subs. & $0.437_{\scriptstyle .006}$ & $0.505_{\scriptstyle .009}$ & $0.336_{\scriptstyle .006}$ & $0.833_{\scriptstyle .005}$ \\
LSTM (Linear), Rnd.\ bins & $0.663_{\scriptstyle .005}$ & $0.664_{\scriptstyle .011}$ & $0.508_{\scriptstyle .007}$ & $0.921_{\scriptstyle .004}$ \\
LSTM (Linear), Rnd.\ win. & $0.667_{\scriptstyle .005}$ & $0.692_{\scriptstyle .010}$ & $0.544_{\scriptstyle .007}$ & $0.929_{\scriptstyle .003}$ \\
Deep BiLSTM & $0.521_{\scriptstyle .005}$ & $0.611_{\scriptstyle .011}$ & $0.416_{\scriptstyle .007}$ & $0.901_{\scriptstyle .003}$ \\
Deep BiLSTM, Rnd.\ subs. & $0.577_{\scriptstyle .005}$ & $0.634_{\scriptstyle .011}$ & $0.460_{\scriptstyle .007}$ & $0.905_{\scriptstyle .004}$ \\
Deep BiLSTM, Rnd.\ bins & $0.759_{\scriptstyle .005}$ & $0.729_{\scriptstyle .010}$ & $0.614_{\scriptstyle .008}$ & $0.945_{\scriptstyle .003}$ \\
Deep BiLSTM, Rnd.\ win. & $0.704_{\scriptstyle .005}$ & $0.721_{\scriptstyle .010}$ & $0.580_{\scriptstyle .008}$ & $0.942_{\scriptstyle .003}$ \\
Deep BiLSTM (GELU) & $0.538_{\scriptstyle .005}$ & $0.579_{\scriptstyle .011}$ & $0.438_{\scriptstyle .007}$ & $0.884_{\scriptstyle .004}$ \\
Deep BiLSTM (GELU), Rnd.\ subs. & $0.647_{\scriptstyle .005}$ & $0.708_{\scriptstyle .011}$ & $0.541_{\scriptstyle .007}$ & $0.923_{\scriptstyle .004}$ \\
Deep BiLSTM (GELU), Rnd.\ bins & $0.716_{\scriptstyle .005}$ & $0.691_{\scriptstyle .010}$ & $0.579_{\scriptstyle .007}$ & $0.932_{\scriptstyle .003}$ \\
Deep BiLSTM (GELU), Rnd.\ win. & $\textbf{0.763}_{\scriptstyle .005}$ & $\textbf{0.733}_{\scriptstyle .010}$ & $\textbf{0.625}_{\scriptstyle .008}$ & $\textbf{0.948}_{\scriptstyle .003}$ \\
Deep BiLSTM (wide head) & $0.587_{\scriptstyle .005}$ & $0.615_{\scriptstyle .009}$ & $0.474_{\scriptstyle .007}$ & $0.900_{\scriptstyle .004}$ \\
Deep BiLSTM (wide head), Rnd.\ subs. & $0.676_{\scriptstyle .005}$ & $0.676_{\scriptstyle .010}$ & $0.538_{\scriptstyle .008}$ & $0.923_{\scriptstyle .004}$ \\
Deep BiLSTM (wide head), Rnd.\ bins & $0.554_{\scriptstyle .005}$ & $0.631_{\scriptstyle .010}$ & $0.445_{\scriptstyle .007}$ & $0.893_{\scriptstyle .004}$ \\
Deep BiLSTM (wide head), Rnd.\ win. & $0.667_{\scriptstyle .005}$ & $0.703_{\scriptstyle .010}$ & $0.545_{\scriptstyle .007}$ & $0.932_{\scriptstyle .004}$ \\
Deep BiLSTM (GELU, wide head) & $0.557_{\scriptstyle .005}$ & $0.604_{\scriptstyle .009}$ & $0.450_{\scriptstyle .007}$ & $0.893_{\scriptstyle .004}$ \\
Deep BiLSTM (GELU, wide), Rnd.\ subs. & $0.644_{\scriptstyle .005}$ & $0.652_{\scriptstyle .011}$ & $0.503_{\scriptstyle .008}$ & $0.913_{\scriptstyle .004}$ \\
Deep BiLSTM (GELU, wide), Rnd.\ bins & $0.740_{\scriptstyle .005}$ & $0.724_{\scriptstyle .011}$ & $0.601_{\scriptstyle .008}$ & $0.938_{\scriptstyle .004}$ \\
Deep BiLSTM (GELU, wide), Rnd.\ win. & $0.569_{\scriptstyle .005}$ & $0.662_{\scriptstyle .010}$ & $0.481_{\scriptstyle .007}$ & $0.915_{\scriptstyle .004}$ \\
\hline
Transformer (Linear) & $0.529_{\scriptstyle .005}$ & $0.583_{\scriptstyle .012}$ & $0.434_{\scriptstyle .008}$ & $0.867_{\scriptstyle .005}$ \\
Transformer (Linear), Rnd.\ subs. & $0.609_{\scriptstyle .005}$ & $0.577_{\scriptstyle .011}$ & $0.453_{\scriptstyle .007}$ & $0.870_{\scriptstyle .005}$ \\
Transformer (Linear), Rnd.\ bins & $0.666_{\scriptstyle .005}$ & $0.621_{\scriptstyle .010}$ & $0.504_{\scriptstyle .008}$ & $0.902_{\scriptstyle .004}$ \\
Transformer (Linear), Rnd.\ win. & $0.559_{\scriptstyle .005}$ & $0.621_{\scriptstyle .011}$ & $0.467_{\scriptstyle .007}$ & $0.890_{\scriptstyle .005}$ \\
Deep Transf. & $0.649_{\scriptstyle .005}$ & $0.588_{\scriptstyle .011}$ & $0.492_{\scriptstyle .008}$ & $0.887_{\scriptstyle .004}$ \\
Deep Transf., Rnd.\ subs. & $0.619_{\scriptstyle .005}$ & $0.647_{\scriptstyle .011}$ & $0.508_{\scriptstyle .008}$ & $0.899_{\scriptstyle .004}$ \\
Deep Transf., Rnd.\ bins & $0.668_{\scriptstyle .005}$ & $0.661_{\scriptstyle .011}$ & $0.539_{\scriptstyle .008}$ & $0.914_{\scriptstyle .004}$ \\
Deep Transf., Rnd.\ win. & $0.628_{\scriptstyle .005}$ & $0.639_{\scriptstyle .012}$ & $0.516_{\scriptstyle .007}$ & $0.905_{\scriptstyle .004}$ \\
Deep Transf.\ (GELU) & $0.641_{\scriptstyle .005}$ & $0.646_{\scriptstyle .011}$ & $0.507_{\scriptstyle .008}$ & $0.904_{\scriptstyle .004}$ \\
Deep Transf.\ (GELU), Rnd.\ subs. & $0.630_{\scriptstyle .005}$ & $0.636_{\scriptstyle .011}$ & $0.515_{\scriptstyle .008}$ & $0.907_{\scriptstyle .004}$ \\
Deep Transf.\ (GELU), Rnd.\ bins & $0.632_{\scriptstyle .005}$ & $0.650_{\scriptstyle .011}$ & $0.515_{\scriptstyle .007}$ & $0.917_{\scriptstyle .003}$ \\
Deep Transf.\ (GELU), Rnd.\ win. & $0.684_{\scriptstyle .005}$ & $0.674_{\scriptstyle .011}$ & $0.543_{\scriptstyle .007}$ & $\textbf{0.923}_{\scriptstyle .003}$ \\
Deep Transf.\ (wide head) & $0.588_{\scriptstyle .006}$ & $0.621_{\scriptstyle .011}$ & $0.488_{\scriptstyle .007}$ & $0.894_{\scriptstyle .004}$ \\
Deep Transf.\ (wide head), Rnd.\ subs. & $0.606_{\scriptstyle .005}$ & $0.676_{\scriptstyle .011}$ & $0.510_{\scriptstyle .007}$ & $0.907_{\scriptstyle .004}$ \\
Deep Transf.\ (wide head), Rnd.\ bins & $\textbf{0.692}_{\scriptstyle .005}$ & $\textbf{0.678}_{\scriptstyle .011}$ & $\textbf{0.556}_{\scriptstyle .008}$ & $0.917_{\scriptstyle .004}$ \\
Deep Transf.\ (wide head), Rnd.\ win. & $0.676_{\scriptstyle .005}$ & $0.676_{\scriptstyle .011}$ & $0.531_{\scriptstyle .007}$ & $0.914_{\scriptstyle .004}$ \\
Deep Transf.\ (GELU, wide) & $0.629_{\scriptstyle .005}$ & $0.622_{\scriptstyle .011}$ & $0.485_{\scriptstyle .007}$ & $0.894_{\scriptstyle .005}$ \\
Deep Transf.\ (GELU, wide), Rnd.\ subs. & $0.607_{\scriptstyle .005}$ & $0.655_{\scriptstyle .011}$ & $0.495_{\scriptstyle .007}$ & $0.904_{\scriptstyle .004}$ \\
Deep Transf.\ (GELU, wide), Rnd.\ bins & $0.655_{\scriptstyle .005}$ & $0.670_{\scriptstyle .011}$ & $0.530_{\scriptstyle .007}$ & $0.914_{\scriptstyle .004}$ \\
Deep Transf.\ (GELU, wide), Rnd.\ win. & $0.661_{\scriptstyle .005}$ & $0.647_{\scriptstyle .010}$ & $0.514_{\scriptstyle .009}$ & $0.912_{\scriptstyle .004}$ \\
\end{tabular}
\end{ruledtabular}
\end{table}

\clearpage
\section{Conclusions}

This work provides the first comprehensive quantitative comparison of feature-based machine learning and end-to-end deep learning on the ZTF StarEmbed dataset. We rigorously measure the performance gap between these paradigms, establishing a clear baseline for future research.

We show that a range of LSTM and Transformer architectures trained on raw light curves consistently underperform models based on physics-motivated features. Data augmentation substantially improves their performance, suggesting that limited labeled data contributes to this gap, although it does not fully explain it. With about 40,000 objects, deep learning models still struggle to generalize, likely due in part to the multivariate, irregular, and morphologically diverse nature of light curves. These findings motivate further investigation of scaling labeled data and self-supervised pretraining on large unlabeled surveys.

We also introduce three novel augmentations, Random Window, Random Bins, and Random Subsampling, that use only real observations. Unlike synthetic methods, they preserve physical integrity while increasing diversity through the natural irregular sampling of astronomical surveys.

We systematically evaluate masked aggregation strategies combining mean, maximum, standard deviation, and final hidden states, instead of simple last-state pooling. This adaptation to irregular, multi-band light curves yields consistent performance gains across both architectures.

Finally, we provide evidence-based justification for moving from task-specific supervised learning to self-supervised foundation models pretrained on massive unlabeled surveys. Our results indicate that classical feature-based methods are approaching saturation, while deep learning remains constrained primarily by data availability, guiding future research directions in astro-informatics.

\begin{acknowledgments}
This research was supported in part through computational resources of high-performance computing (HPC) facilities at HSE University \cite{19}. 
\end{acknowledgments}

\section*{FUNDING}
The article was prepared within the framework of the project “Mirror Laboratories” HSE University. Support was provided by Schmidt Sciences, LLC, for K. Malanchev.

\section*{CONFLICT OF INTEREST}
The authors of this work declare that they have no conflicts of interest.

\end{document}